# SEMI-FND: Stacked Ensemble Based Multimodal Inference For Faster Fake News Detection

Prabhav Singh*, Ridam Srivastava*, K.P.S. Rana, Vineet Kumar
Instrumentation and Control Engineering Department
Netaji Subhas University of Technology, Sector-3, Dwarka, New Delhi - 110078, India
Phone: +91-11-25099050, Fax: + 91-11-2509905022, Website: www.nsut.ac.in
Emails: prabhavs.ic18@nsut.ac.in, ridams.ic18@nsut.ac.in, kpsrana@nsut.ac.in, vineet.kumar@nsut.ac.in

**ABSTRACT**

Fake News Detection (FND) is an essential field in natural language processing that aims to identify and check the truthfulness of major claims in a news article to decide the news veracity. FND finds its uses in preventing social, political and national damage caused due to misrepresentation of facts which may harm a certain section of society. Further, with the explosive rise in fake news dissemination over social media, including images and text, it has become imperative to identify fake news faster and more accurately. To solve this problem, this work investigates a novel multimodal stacked ensemble-based approach (SEMI-FND) to fake news detection. Focus is also kept on ensuring faster performance with fewer parameters. Moreover, to improve multimodal performance, a deep unimodal analysis is done on the image modality to identify NasNet Mobile as the most appropriate model for the task. For text, an ensemble of BERT and ELECTRA is used. The approach was evaluated on two datasets – Twitter MediaEval and Weibo Corpus. The suggested framework offered accuracies of 85.80% and 86.83% on the Twitter and Weibo datasets respectively. These reported metrics are found to be superior when compared to similar recent works. Further, we also report a reduction in the number of parameters used in training when compared to recent relevant works. SEMI-FND offers an overall parameter reduction of at least 20% with unimodal parametric reduction on text being 60%. Therefore, based on the investigations presented, it is concluded that the application of a stacked ensembling significantly improves FND over other approaches while also improving speed.

* Denotes those authors have equal contribution.

## 1. INTRODUCTION

The advent of the World Wide Web in the 20th century has given a new meaning to the term "Fake News". One might assume that the term has always carried the same meaning in society. However, fake news today is not what it was before the arrival of the internet. Before the mid-1990s, fake news either referred to satire and intentional sensationalism, which was aimed at increasing readership and gathering attention or referred to the news propagandas aimed at discrediting organizations or ideologies [1] [2]. Today, the most common definition of fake news [3] is - "news articles that are intentionally and verifiably false."
Modern fake news is aimed at spreading false information with harmful intent and is sometimes generated and propagated by hostile foreign actors – as seen in multiple cases in recent times. According to Zhang et al. [4], modern fake news is characterized by differences in three features – Volume, Veracity and, Velocity. The Volume of fake news has seen a substantial increase in recent times due to a decrease in the verification process. While newspapers and print media involved layers of verification, the internet has allowed a larger user base to publish fake news and stories [5]. This is shown in Fig. 1, which displays the interest in the term 'fake news' as obtained from Google Trends [6]. Second, the Variety of fake news has increased to incorporate fake reviews, satires, false advertisements, misinformation, and disinformation. Finally, the Velocity of the dissemination of fake news has increased manifolds due to the advent of social media [4].

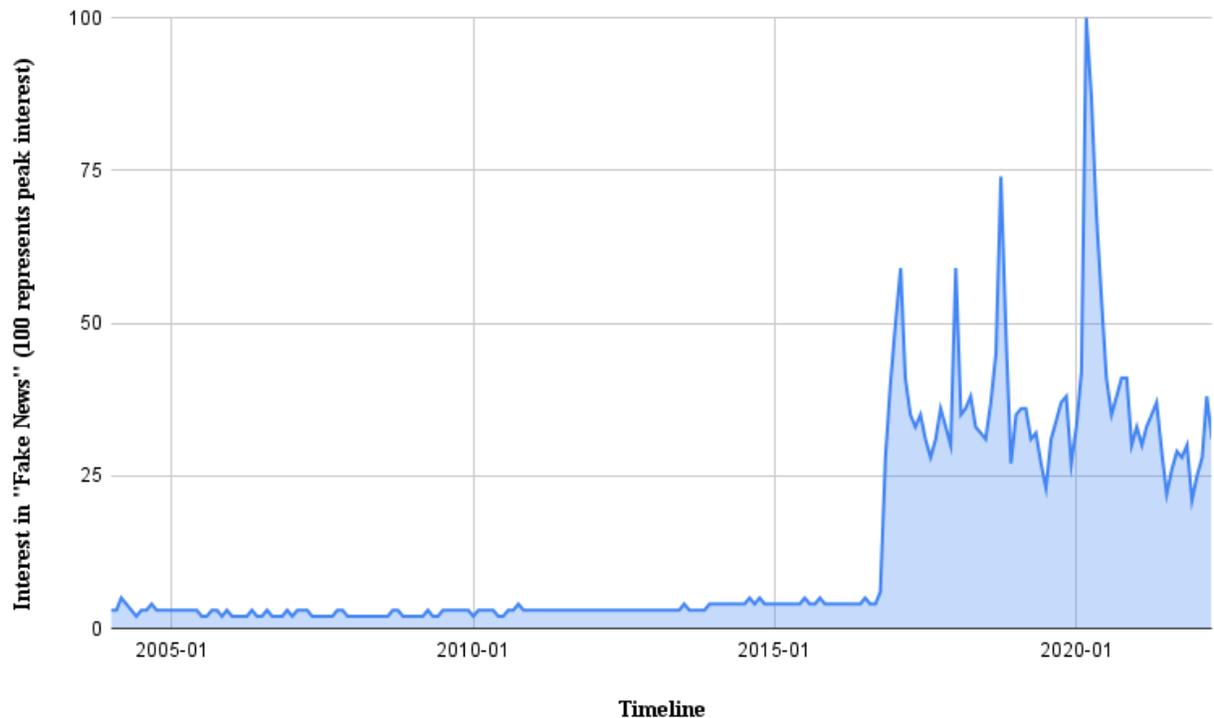

**Fig. 1.** Interest in the term – 'Fake News'

This exponential rise in the amount of fake news on the internet, has invariably led to an increase in major events that have triggered the recent interest of researchers on this problem. According to Balmas [7], the major contributors of this problem have been the rise of social media platforms like WhatsApp, Facebook, Instagram and, most majorly, Twitter. A major example of the same is the effect of fake news during the 2016 US Presidential Elections. According to [8], this was linked to an increase in polarization and partisan conflict during the election process. Furthermore, their analysis of 171 million Tweets during the election process showed that 25% of the tweets were extremely biased or fake.

This issue of misinformation has also become a pivotal factor in the status quo due to the COVID-19 pandemic. On social media, false COVID-19 cures such as using bleach as an injection have proliferated [9], as have false conspiracy theories that the virus was created in a Wuhan lab [10], amongst many other theories. This dissemination of misinformation becomes a larger issue due to its impact on human life, especially in third-world countries with inadequate knowledge and resources required for dealing with such events. In Bangladesh for instance, rumours related to Ramu Violence (2012) [11] lead to the disruption of social stability, while in India fake news spread by extremists has resulted in violence against minorities, specifically through incidents related to cow e-vigilantes. An example of tweets related to these incidents is shown in Fig. 2.

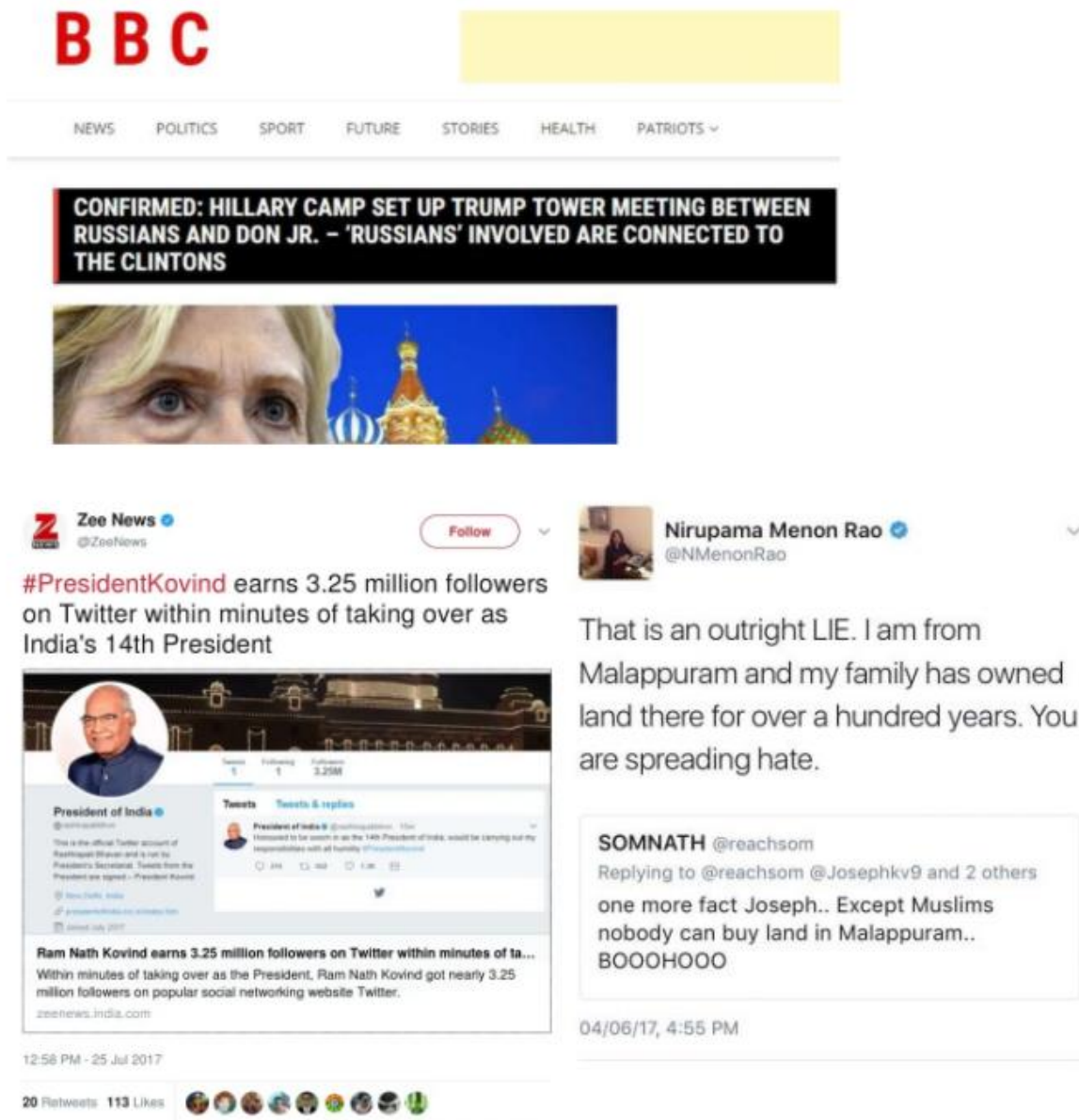

**Fig. 2.** In clockwise order: (i) Fake news during the US Presidential Elections (2016), (ii) Fake news aimed at inciting communal riots, (iii) Fake news regarding the social presence of the Indian President.

From the above discussion, it can be concluded that it is imperative to research more robust and accurate models for fake news detection. Ideally, a fake news detection system should be one that can fulfill the following needs sufficiently well:
1. Accurately detect fake news while also being robust enough to counter the false-positive trap which can lead to fake news being classified as real news.
2. Increase processing speed to counter the surge in the amount of fake news on social media platforms.
3. Ensure performance on all types of data circulated online – ranging from text to images, GIFs, and videos.

Since the surge of fake news in 2014, multiple works have aimed to solve this issue by employing a variety of methods. Early methods [12,13] attempted to focus only on statistical and word-based features for FND. For example, Rashkin et al. [14] employed the use of adverbs and their types to identify fake news from text. On the other hand, Nakashole and Mitchell [12] attempted to train traditional models using the text and its source as the input features. However, recent works [15,16] have recognized the importance of including more modalities while increasing the focus on utilizing context-aware techniques. This is because today, purely textual based methods have become redundant due to three major reasons:
1. The primary factor remains the short length of most modern news mediums. According to a survey by [17], Twitter remains one of the major sources of fake news, which has a constraint of 150 characters. This can lead to difficulties in fine-tuning machine learning models to verify the veracity of tweets.
2. Moreover, the increase in intentional and targeted fake news has caused the writing style of most news articles to be very similar to genuine news. This reduced separation between the two also increases the complexity of the task.
3. Finally, with an increase in the popularity of interactive content (GIFs, videos, and images), most fake news is disseminated majorly through other modalities. While text remains an essential component, without the support of other modalities, it has become almost impossible to make the distinction with acceptable accuracies [17].

Due to these reasons, multimodal approaches to FND have seen an increased interest in recent times. Shim et al. [17] employed the use of links present in tweets and articles as a separate modality to increase the contextual knowledge of their classifier. They also showed an increase in performance when compared to conventional text-only methods applied to the same data. Further, Singhal et al. [18] reported a substantial increase in performance with the inclusion of images as a modality for FND tasks.  However, none of these works have attempted to establish a comparative analysis of image models for FND tasks. Further, most works have only explored the use of the popular BERT text model for the processing of lexical features. Furthermore, none of the works have explored the application of ensemble-based decision-making for the text modality. It has been previously shown that the ensembling of text models can lead to improved performance [19–21]. Though some works have recently started investigating methods to reduce the processing speed of FND systems [19,20] there has overall been very little focus and no concrete conclusions as of now.

Drawing motivation from the above background, the proposed work presents a novel stacked ensemble-based architecture for multimodal fake news detection by employing decision level fusion of modalities. Further the work presents a comparative analysis of 18 pre-trained image models to select the most accurate model for image processing while keeping the number of parameters as low as possible to improve speed.

Furthermore, the work introduces an ensemble of two deep learning architectures – ELECTRA and BERT for textual features. The ensemble is merged at the decision level using equal weightage averaging. Finally, image and text modalities are also merged at the decision level with the use of equal weightage averaging. This paper thus proposes a novel extractive FND method that focuses on both images and text modalities while aiming to improve the overall speed of the task. To evaluate its performance, the proposed method is tested on two datasets: Twitter MediaEval Dataset (2016) [22] and Weibo Fake News Corpus. The significant contributions of the paper can be summarized as follows:

1. A deep learning based stacked ensemble multimodal architecture for FND is successfully presented and tested.
2. A comparative study of pretrained image models is carried out on two datasets to select the most accurate yet fast model.
3. The use of an ensemble of two pretrained text models: BERT and ELECTRA are proposed and successfully tested.
4. The work also reports a 20% reduction in image-based training parameters and a 122% reduction in text-based training parameters when compared to recent works [23][24]. Overall, the presented work is shown to be faster than recent counterparts while achieving superior performance.

The paper is organized as follows. Following the introduction in Section 1, Section 2 presents a brief survey of related works in the field. Section 3 describes the developed approach and proposed pipeline along with the above-mentioned comparative analysis of image models. It also includes a detailed overview of the datasets used and a brief study of the evaluation metrics. Section 4 summarizes the findings and provides a comparison with recent similar works. A comparative analysis with respect to the runtime of proposed and recent models is also presented. Finally, Section 5 brings the study to a close with some potential future ideas.

## 2. RELATED WORKS

This section is further divided into two subsections. Subsection 2.1 presents works that utilize a traditional approach to FND systems. These include both statistical methods and supervised machine learning based methods. Subsection 2.2 presents works that utilize deep learning to classify news as fake or real. These also include multimodal systems for FND. Finally, inferences and research gaps are drawn and presented.

### 2.1 Traditional Methods

Since the formulation of the problem of fake news detection, one of the most common approaches proposed has been that of exploiting linguistic features that capture different writing styles and sensational headlines. These features are extracted from the text in terms of document organizations from different levels, such as characters, words, sentences, and documents, and can uncover various relationships that aid in the detection of fake news [25].

Suggesting one such correlation between the objectivity of language and trustworthiness of sources, Nakashole et al. [26] introduced a believability computational model, FactChecker. Based on their initial analysis, the use of neutral and impartial language was found to be more closely related to trustworthy sources than a speculative and opinionated style of writing. This hypothesis formed the basis for calculating believability scores, along with the incorporation of the influence of co-mentions. Corroborating on these findings and investigating further, Rashkin et al. [14] performed a deeper linguistic comparison of different types of fake news, i.e., propaganda, satire and hoaxes and for varying levels of truth. They reported uncertainty and vagueness to be a characteristic of false information, along with the use of subjective, superlatives, and modal adverbs. Khurana [27] on the other hand, not only took into consideration the kind of words used in the text but also employed sentiment analysis to capture the entire emotion of the text while using POS Tagging and n-grams to obtain insights into the semantics and syntax of a statement. These extracted features were tested on several classifiers, with logistic regression yielding the best accuracy of 50.16% on the LIAR dataset. Leveraging logistic regression further, Tacchini et al. [28] tested whether the set of users interacting with news posts on social networking sites can be used to ascertain if the news is fake. While the study reported promising results with logistic regression, the Boolean label crowdsourcing algorithm achieved superior results with more than 99% accuracy on a self-created dataset. Traditional classifiers have also been tested, with results varying primarily in the overall framework and feature extracted. Granik et al. [29] for instance, reported a 74% accuracy with the Naive Bayes classifier on the Buzzfeed News dataset, employing a simple classification approach without any additional attributes.

In [30], Katsaros et al. evaluated representatives from eight well-known families of classification algorithms and concluded a space with hundred dimensions to be of adequate dimensionality to capture the needed text features with high accuracy. Moreover, Ahmad et al. in [31], evaluated individual learning classifiers against various ensemble models used as voting classifiers. These ensemble models included a combination of logistic regression, random forest, and KNN, and another of logistic regression, linear SVM, and classification and regression trees. The study reported that ensemble learners show an overall better score on all performance metrics as compared to the individual learners, serving as a motivation to adopt the ensemble approach in this work. Lastly, the use of Rhetorical Structure Theory in an analytic framework to identify systematic differences between false and truthful stories was explored by Rubin et al. [32]. The authors utilized a Vector Space Model (VSM) to assess each story's position in a multi-dimensional RST space with respect to its distance to truth and treated deceptive centers as a measure of the story's degree of truthfulness.

## 2.1 Deep-Learning and Multimodal Methods

Most recent works on fake news detection have concentrated on leveraging neural networks for this task and have reported significant improvements over the baseline results. Singhania et al. [33] proposed a three-level hierarchical attention network, one each for words, sentences, and the headline to construct a news vector, by processing an article in a hierarchical bottom-up manner. Because of the three layers of attention, the model gave differential importance to all parts of an article, achieving an accuracy of 96.77% on a large real-world dataset. Using a deep convolutional neural network (CNN), Kaliyar et al. [34] used the model to automatically learn the discriminatory features through multiple hidden layers built in the network. By extracting several features at each layer and employing GloVe as a pre-trained word embedding, they reported an improvement of 5% over state-of-the-art results on the Kaggle news dataset. GloVe embeddings have also been found to be more useful in providing vector projections that reflect relations, similarities, dissimilarities with other words than the traditional Bag of Words, as reported by Aggarwal et al. in [35]. The study concentrated on solely relying on text processing without the context of the history and credibility of the author or source and achieved benchmark results due to word embeddings complementing the CNN and recurrent neural networks (RNN) based model.

The utility of CNN and RNN in fake news prediction has further been investigated by Bahad et al [36]. In [36], the authors evaluated and compared the accuracy of a Bi-directional LSTM-RNN model with CNN, vanilla RNN, and unidirectional LSTM-RNN. While CNN was found to perform better in extracting local and position-invariant features, LSTM-RNN was reported to be more useful for a long-range semantic dependency-based classification. Overall, Bi-directional LSTM-RNN model were reported to be significantly more effective than unidirectional models. Exploring the efficacy of semi-supervised approaches, Benamira et al. [37] tested a simple nearest neighbor graph among articles based on word embedding similarities, accompanied by graph neural networks for classification. They achieved a performance improvement of up to 3% with only 10% of the labelled data, while also reducing the standard deviation in results and taking less time for processing. Their work provided a basis for semi-supervised content-based detection methods.

Recently there has been an increased interest in exploiting visual features along with textual context, as is explored in this work. Attempting to capture the hidden patterns in the words and images used in fake news, authors in [38] extracted latent features via multiple convolutional layers. Their framework projected explicit and latent features into a unified feature space and was trained on both text and image inputs simultaneously. They achieved a precision of 92.20% on a dataset focused on news regarding the American presidential elections. Unlike the event-specific features learnt in [38], Wang et al. [23] proposed an end-to-end framework of adversarial neural networks to derive event-invariant features and thus facilitate the detection of fake news on newly arrived events. The work combines a multi-modal feature extractor for linguistic and visual features, a fake news detector to learn the discriminable representations and an event discriminator for removing features specific to an event while keeping the shared features among different events.

Comparable to the methodology adopted in this work, SpotFake [39] is a multimodal framework leveraging language models to learn text features and pre-trained ImageNet models for extracting features from images. Combining word embeddings from BERT and feature representations from VGG-19, the two modalities are concatenated to form the news feature vector. Although the work employs computationally expensive models, it outperforms the baselines by a margin of 6% accuracy on average. Besides images, behavior of contributing parties in fake news propagation has also been explored [40]. Ruchansky et al. [40] integrated the temporal pattern of user activity on a given article with the source characteristic based on the behavior of users to perform classification and reported an accuracy of 89.20% on the Twitter Fake News dataset. Alrubaian et al. [41] proposed a framework for credibility analysis to assess information credibility on Twitter as a means of preventing the proliferation of malicious information. Their proposed framework integrated four components - a reputation-based component to filter neglected information, a credibility classifier engine that distinguishes between credible and non-credible content, a user experience component that yields ratings of Twitter-user expertise on a specific topic, and a feature-ranking algorithm for selecting the best features based on their relative importance. Using conventional classifiers like random forests and naive Bayes for these components, the work achieved a significant balance between recall and precision on two datasets from 489,330 unique Twitter accounts. Along similar lines, Shu et al. [42] investigated the importance of modelling user-user relations and news-user relations to capture effective feature representations. The authors explored the correlations of publisher bias, news stance, and relevant user engagements, in a Tri-Relationship Fake News detection framework. The framework extracted effective features from news publishers and user engagements separately, while also capturing the interrelationship, leading to good detection performance in the early stages of news dissemination. In [43], Aphiwongsophon et el. combined the text from tweets with metadata attributes such as verification status, followers count, number of hashtags and more. Naive Bayes and SVM classifiers were trained on the above feature set and the results were found comparable to that of simple neural networks, indicating that combining linguistic features with raw data attributes can be beneficial. Moving forward on this path [44], Gupta et al. made use of sub-word level embeddings from FastText, contextual embeddings from BERT and TF-IDF based ranking for the construction of a Bag of Words feature vector, all of which were combined with the user metadata to form a single feature vector. Their study reported an F1 score of 0.93 on the English Fake news dataset, in agreement with the findings of [44].

Based on the above survey, the following three inferences were drawn.
1. The benefit of ensembling is clearly established when implemented in supervised ML techniques.

2. Deep learning networks and specifically, pre-trained word embeddings have been shown to be superior when compared to feature-based classifiers.
3. Ensembling of pre-trained word embeddings has not been utilized for FND tasks.
4. Decision level fusion of modalities has been shown to be better than feature level fusion for FND tasks.
5. Very few works have focused on parameter reduction and on improving training time.

Motivated by the above inferences, this work introduces a novel stacked ensemble based multimodal architecture for faster fake news detection. To improve performance on image modality while keeping processing time low, unimodal analysis is carried out on 18 models. Pre-trained word embedding models are selected to create the ensemble for text analysis such that the number of parameters remain low while maintaining good accuracy. Decision level fusion is used to combine modalities and to combine stacked text models. The detailed methodology is further elaborated upon in Section 4.

## 3. BRIEF DESCRIPTION OF USED DATASETS

### 3.1 Twitter MediaEval Dataset[22]

The Twitter MediaEval Dataset [22] was curated as part of the '2015 Verifying Multimedia Use Challenge' conducted by MediaEval. The task involved identifying whether the multimedia items accompanying a tweet reflect the reality in the way purported by the tweet. The dataset comprises 17,000 unique tweets and their associated images, collected around several widely known events or news stories. The dataset is divided into two parts: the development set (9000 false news tweets, 6000 actual news tweets) and the test set (9000 fake news tweets, 6000 real news tweets) (2000 tweets). By cross-checking online sources such as articles and blogs, these tweets have been manually verified and classified into 'real' and 'fake' data points. A tweet is marked to be real if the corresponding image represents the event that the tweet refers to, while a fake tweet contains images that are not relevant to the event described in the tweet. For usage in this work, all datapoints with only unimodal information were removed from the dataset. Further, all datapoints with videos and GIFs were not considered. After these modifications, details of the corpus are shown in Table 1.

### 3.2 Weibo Dataset [45]

The Weibo NER dataset is a Chinese Named Entity Recognition dataset containing false rumor posts from May 2012 to January 2016 collected from the social network, Sina Weibo. This multi-media dataset containing images, Graphics Interchange Format (GIF) and other media has all its posts verified by the official rumour debunking system of Weibo [46]. Though Weibo follows the microblogging model of Twitter, there are some significant distinctions between the two [45]. Firstly, some linguistic aspects usually studied for English tweets, such as case sensitivity of English words, repeated letters, and word lengthening, do not apply to the Chinese language used in Weibo. Moreover, the types of microblogs retweeted differ from those on Twitter. The datasets used in this work are thus varied and contribute to testing the robustness of the proposed framework.

**Table 1.** Summary of datasets used

| Dataset | Training | | Testing | |
|---|---|---|---|---|
| | Real | Fake | Real | Fake |
| Twitter MediaEval | 5008 | 7032 | 1217 | 2564 |
| Weibo | 4274 | 4274 | 475 | 475 |

## 4. PROPOSED METHODOLOGY

This section describes the methodology adopted in the proposed work. This section begins by providing a detailed overview of the suggested framework in Fig 3. Subsection 4.1 introduces the preprocessing steps applied to visual and textual features. Following that, Subsection 4.2 explains the procedure used for unimodal model analysis for the image modality. It also includes the results of the analysis which in turn represent the final model used for images. Subsection 4.3 introduces the text embedding models employed, namely BERT and ELECTRA. The ensembling of the text models along with the decision level fusion of text and image modalities is explained in Subsection 4.4. Finally, Subsection 4.5 introduces the metrics used for evaluating the efficacy of the approach

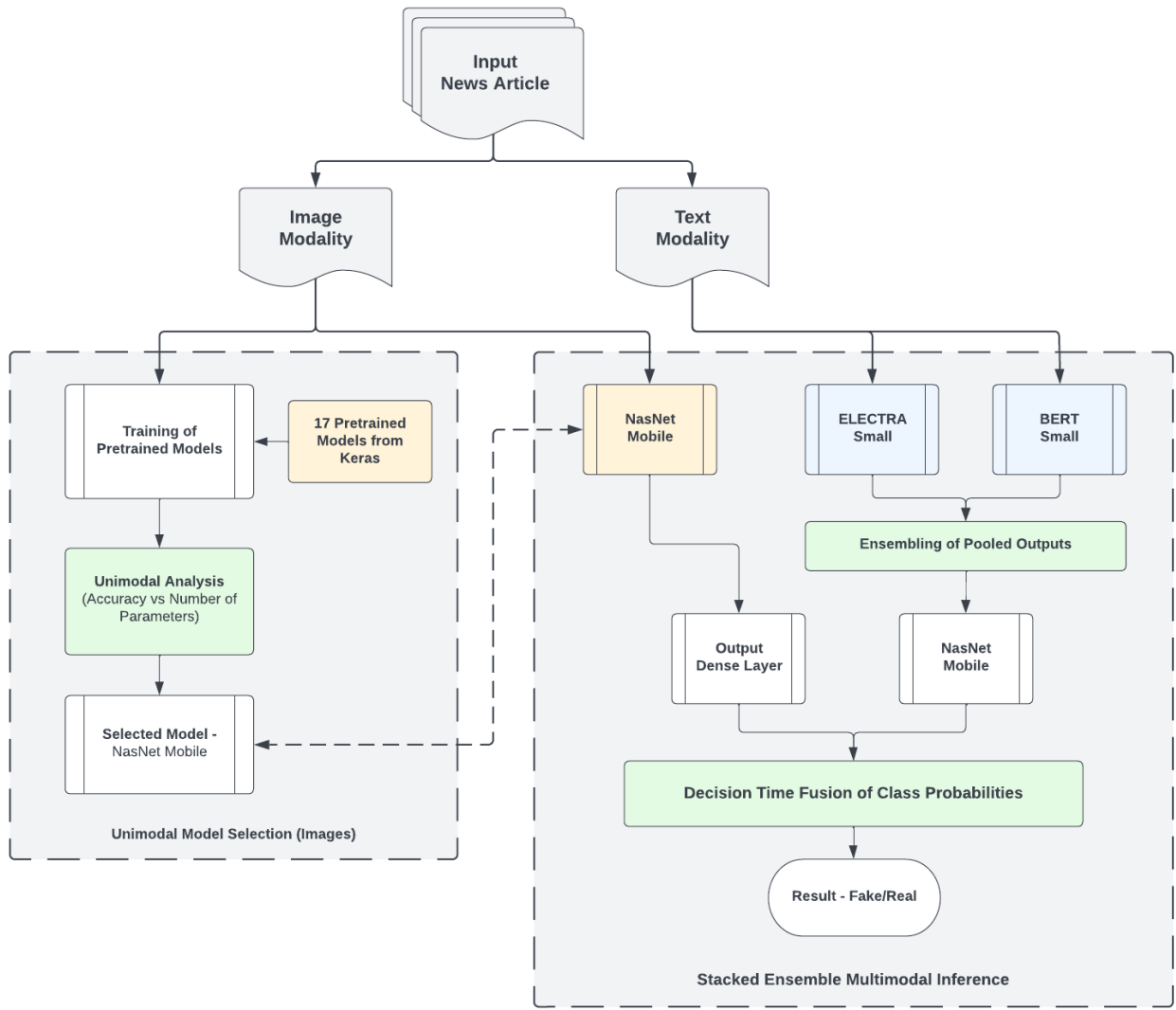

**Fig. 3.** Detailed description of the proposed methodology.

## 4.1 Preprocessing of Images and Text

Prior to unimodal model selection for images and application of the pre-trained text embeddings on the text, various preprocessing steps were applied to the data to improve results. The preprocessing of text was kept as minimal as possible due to the already low character count of input data. The preprocessing steps applied to the images were:

1. ***Removal of GIFs and Videos:*** In line with recent proposed works [23,39], both datasets were scanned and all datapoints including GIFs and videos as the only supporting modality were removed. This was done to keep the required processing power at a minimum and to generalize the training procedure.

2. **Resizing of Image:** Since all pre-trained models provided by Keras [47] have a standard input image dimensionality of (224, 254, 3), all images from both datasets were resized to these dimensions and stored.
3. **Image Normalization & Noise Addition:** Post resizing, normalization of all images was carried out. This was done by applying Min-Max Normalization on the inter-cubic distances of all images. To improve training and to prevent overfitting, given the low number of unique images in both datasets, Gaussian Noise [48] was added to all images artificially. The probability density function ($\sigma$) of a Gaussian random variable ($z$) is given in Eq. 1.

$$\rho_g(z) = \frac{1}{\sigma\sqrt{2\pi}} \cdot e^{-\frac{(z-\mu)^2}{2\sigma^2}} \tag{1}$$

*where,*
*μ is the mean grey value;*
*σ is the standard deviation.*

As mentioned before, minimal preprocessing steps were applied to the text. The only applied step included removal of URLs and unwanted characters. Both datasets, specifically the Twitter MediaEval corpus, contain a large number of tweets. These usually have a high frequency of unwanted text such as URLs, email addresses, and hyperlinks. To prevent the effect of such characters on the output, regex expressions were employed to remove URLs and hyperlinks. Post the application of these preprocessing steps, the data was fed to the proposed architecture for inference.

**4.2 Model Selection for the Image Modality**

After preprocessing the images, the training and validation splits of the Twitter MediaEval Dataset and the Weibo Fake News Corpus were used to select the optimal model for image modality inference. For the Twitter dataset, only the development (training) dataset was used with a 75-25% split to ensure no overlap with the test set. However, for the Weibo dataset, no such split is provided. Hence, a 7:1:2 split was used in line with the most popular works on the dataset [23,24]. Since the focus of the work is to devise a multimodal method for FND, all datapoints with only one modality present were removed from consideration. Post these steps, the total number of images on which analysis was carried out was 10,024: 512 images from the Twitter Dataset and 9,528 images from the Weibo Corpus. The analysis was carried out on both datasets and the results from both were averaged to arrive at the final results.

For the selection of the most suited image processing architecture, 17 pre-trained models were obtained from the Keras library. Table 2 summarizes the parameter count and other features of the 17 models. It should be noted that NasNet Large [49] was not used due to a different input shape than all other models. Further, only two variations of the EfficientNet Model [50] were used due to processing limitations. It should also be noted that prior to this work, VGG16 and VGG19 [51] remained the most used models.

**Table 2.** Summary of models used for Image Analysis and Model Selection

| Name | Size (MB) | No. Of Parameters (Millions) | Depth |
|---|---|---|---|
| Xception | 88 | 22.90 | 81 |
| MobileNet | 16 | 4.30 | 55 |
| MobileNet V2 | 14 | 3.50 | 105 |
| MobileNet V3 Large | NA | 5.40 | 217 |
| ResNet101 | 171 | 44.70 | 209 |
| ResNet 101 V2 | 171 | 44.70 | 205 |
| ResNet 152 V2 | 232 | 60.40 | 307 |
| ResNet50 V2 | 98 | 25.60 | 103 |
| InceptionResNetV2 | 215 | 55.90 | 449 |
| EfficientNetB0 | 29 | 5.30 | 132 |
| EfficientNetB7 | 256 | 66.70 | 438 |
| VGG 16 | 528 | 138.40 | 16 |
| VGG 19 | 549 | 143.70 | 19 |
| NASNet Mobile | 23 | 4.30 | 389 |
| DenseNet 121 | 33 | 8.10 | 242 |
| DenseNet 169 | 57 | 14.30 | 338 |
| DenseNet 201 | 88 | 20.20 | 402 |

Training and validation methodologies were kept the same for all models. Input dimensions were kept as (224, 224, 3) as discussed in subsection 4.1 All models were trained and validated on Google Collaboratory Pro [52] on an NVIDIA Tesla T4 GPU. The pooling layers were kept as 'avg' for all models and all layers were frozen. Training was done in a batch size of 8 for 3 epochs and categorical cross-entropy was used as the loss function. For a comparative analysis, the validation accuracy after 5 epochs was considered against the number of parameters in the model. Finally, these results were averaged for both datasets and were used to select the most appropriate model for image inference in FND. Fig 4. shows the average results for both datasets while Table 3 presents the results of the top 5 models.

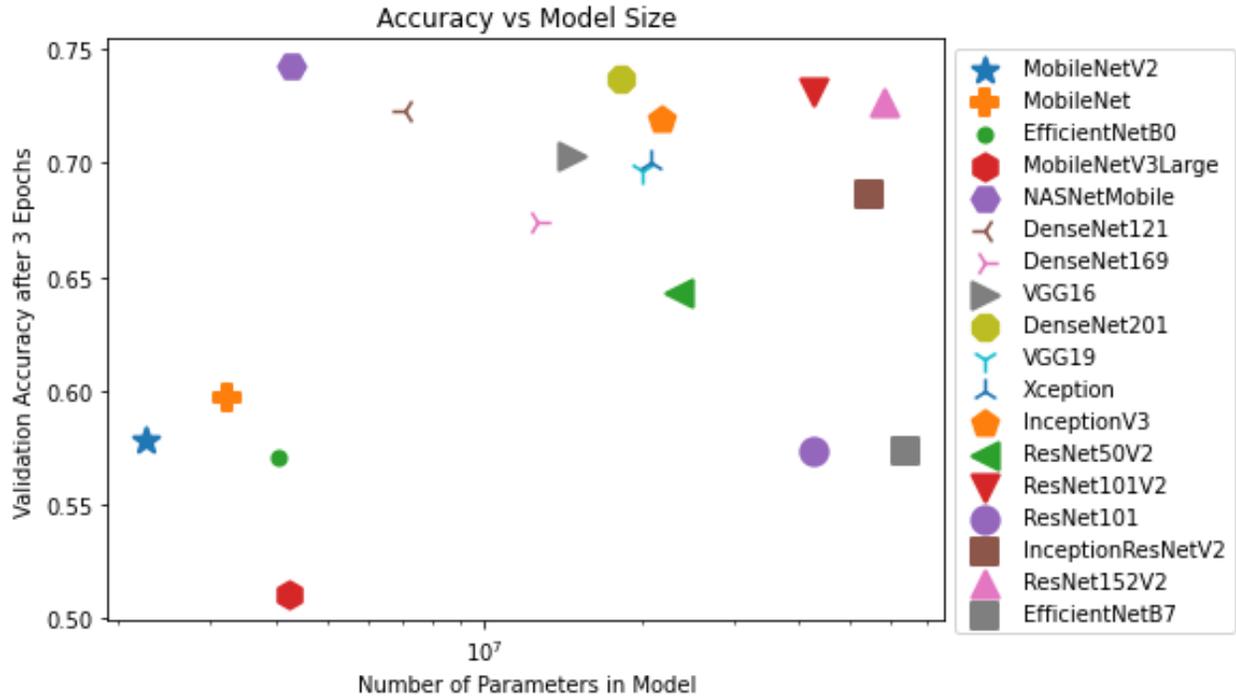

**Fig. 4.** Results of image model analysis on two datasets.

**Table 3.** Results of top 5 models

| Name of Model | Number of Parameters | Validation Accuracy (%) After 3 Epochs |
|---|---|---|
| **NasNet Mobile** | **4,269,716** | **74.31** |
| DenseNet 201 | 18,321,984 | 73.67 |
| ResNet 101 V2 | 42,626,560 | 73.13 |
| ResNet 152 V2 | 58,331,648 | 72.67 |
| DenseNet 121 | 7,037,504 | 72.31 |

It can be clearly seen from the obtained results that NasNet Mobile outperform all other models by a very large margin. While the validation accuracy of the top 5 models remains within 2% of each other, using NasNet Mobile results in a parameter reduction of 76.44% when compared to the second-best model. It is further noted that the model achieves an improvement of 4% over the more commonly used VGG16 architecture while reducing parameters by 70%. For these reasons, NasNet Mobile was finalized as the architecture used for image inference in SEMI-FND.

The NasNet Mobile architecture is based on the Neural Architecture Search (NAS) framework proposed by [53]. In this model, the overall architectures of the convolutional nets are manually predetermined. The model is characterized by repeating convolutional cells where each cell has the same architecture, but different weights. These convolutional cells are in turn of two types, namely, Normal Cells and Reduction Cells. Normal cells return a feature map of the same dimensions as the input feature map, while Reduction Cells return a feature map with the height and width reduced by a factor of two. The structures of these Normal and Reduction cells are searched by a controller RNN which recursively predicts the structure based on the provided hidden initial states. This is formulated as a reinforcement learning problem, with the resulting accuracies used to update the controller so that it generates better architectures over time.

**4.3 Text Embedding Models: BERT and ELECTRA**

In order to leverage information from textual data, vector representations of words were computed using BERT and ELECTRA. Bidirectional Encoder Representations from Transformers, or BERT is a deeply bidirectional model that provides contextualized word embeddings by effectively capturing information from both the right and left context of the input tokens. This is achieved through Masked Language Model (MLM), wherein instead of predicting the next word in a sequence like traditional language models, the discriminator in the model is trained to predict a missing word from within the sequence itself [54]. Around 15% of the words are randomly masked i.e., replaced by [MASK] tokens to indicate the missing word that has to be predicted, enabling the model to understand the relationship between words. In addition, BERT is also trained on the task of Next Sentence Prediction (NSP) to gain an understanding of the relationship between sentences. This effective pre-training approach of MLM and NSP has proved to be superlative in comparison to traditional NLP models with several recent works leveraging the model for NLP tasks. The improved performance, however, comes at great computational costs, with even the small BERT version used in this work requiring 30 million parameters to be trained. To improve the training time required and capture the context of the sentences from a different perspective, the lesser investigated ELCTRA model is also explored in this work. While the underlying architecture and most hyperparameters of ELECTRA are same as BERT, a different training approach is followed for the discriminator. Instead of masking a random selection of input tokens, ELECTRA replaces random tokens with plausible alternatives produced by a generator [55]. The task of the discriminator is then to distinguish between the real and false data, allowing the model to learn from all the input tokens instead of focusing on solely the [MASK] tokens. Both the discriminator and generator consist of an encoder which maps the sequence on input tokens $(x) = [x_1, ..., x_n]$ into a sequence of contextualized vector representations $h(x) = [h_1, ..., h_n]$. For a given position t, where $x(t) = [MASK]$, the generator outputs a probability for generating a particular token with a SoftMax layer given by:

$$p_G(x_t : x) = \frac{\exp(e(x_t)^T h_G(x)_t)}{\sum_{x'} \exp(e(x')^T h_G(x)_t)} \tag{2}$$

where, e denotes token embeddings.

In essence, for a given position t, the discriminator predicts whether the token x(t) comes from the data or the generator distribution i.e., whether the token is "real" or not. This approach to pre-training makes the model computationally more efficient, outperforming BERT base by 5 points on the GLUE benchmark dataset with only 1/20th the parameters of the latter. While ELECTRA-small further reduces the training time, its accuracy reported on benchmark datasets such as SQuAD and GLUE is lesser than that of BERT. Therefore, to leverage the benefits of both models and take into account both training approaches, the presented work utilizes the small BERT model V2 [56] for embedding textual information into a 768-dimensional vector, along with the ELECTRA-small ++ model V2 [57] that produces embeddings of 256 dimensions. By ensembling the outputs of these two models, the overall accuracy for lexical features is greatly improved.

## 4.4 Ensembling of Text Models and NasNet Mobile

As mentioned before, the proposed method employs stacked ensembling of the two text models to capture relevant information from both sets of pooled outputs. Fig. 5 shows the detailed procedure by which this step is achieved. Per subsection 4.3, BERT (Small) and ELECTRA (Small++) were selected as the two text embedding models for the task. Specifically, the small BERT has 12 hidden layers, a hidden size of 768 and 12 attention heads. Both models are obtained via the TFHub library [58].

To obtain the class probabilities from text representation, first the preprocessed text was fed through the text embedding models to obtain the pooled and sequence outputs. For this implementation, only the pooled outputs were used to reduce computation power and increase the speed of the architecture. The outputs were then run through a dense layer of 128 units. This was done for two reasons:
1. The output of the BERT architecture is of the shape (Batch_Size * 768) while that of the ELECTRA architecture is (Batch_Size * 256). To ensure uniform contribution of both text representations, the method reduces both outputs to 128 dimensions.
2. This further contributes to improving the speed of the architecture as well.

After obtaining the final text representation of both models, stacking was performed by concatenating outputs along the length. This resulted in a text representation of shape (Batch_Size * 256). Finally, to arrive at the class probabilities, the concatenated output was fed through a deep neural architecture composed of dense and dropout layers as shown in Fig. 5. A sigmoid layer was used to arrive at the output which was stored for further concatenation with the image modality.

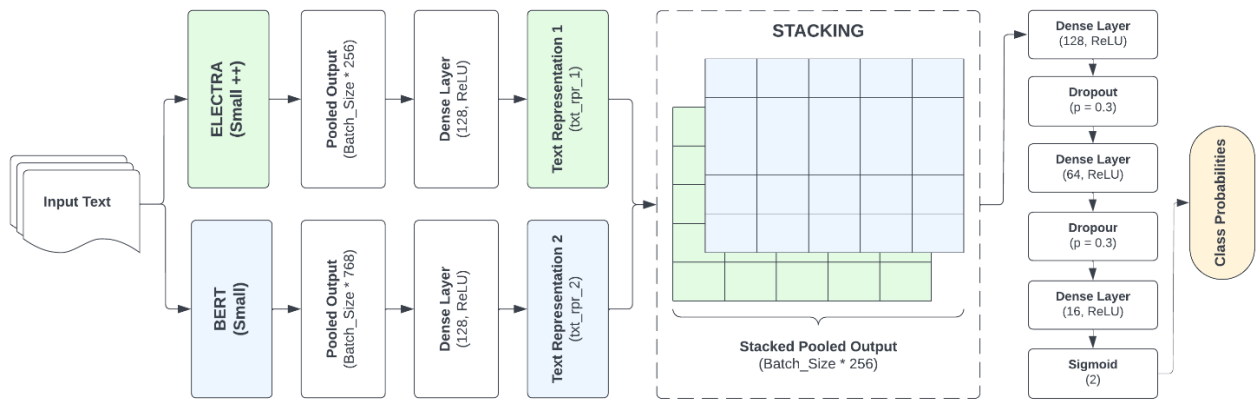

**Fig. 5.** Stacked Ensembling of BERT and ELECTRA models.

After obtaining the class probabilities for text, the same procedure is applied to images with the help of NasNet Mobile. As shown in Fig. 6, the preprocessed images were fed through the NasNet Mobile architecture to obtain the output representation. These output representations were then flattened and fed through a neural architecture to arrive at the image modality class probabilities. Finally, decision level fusion of both modalities was done by averaging the probabilities to arrive at the final result for each class (Fig. 6).

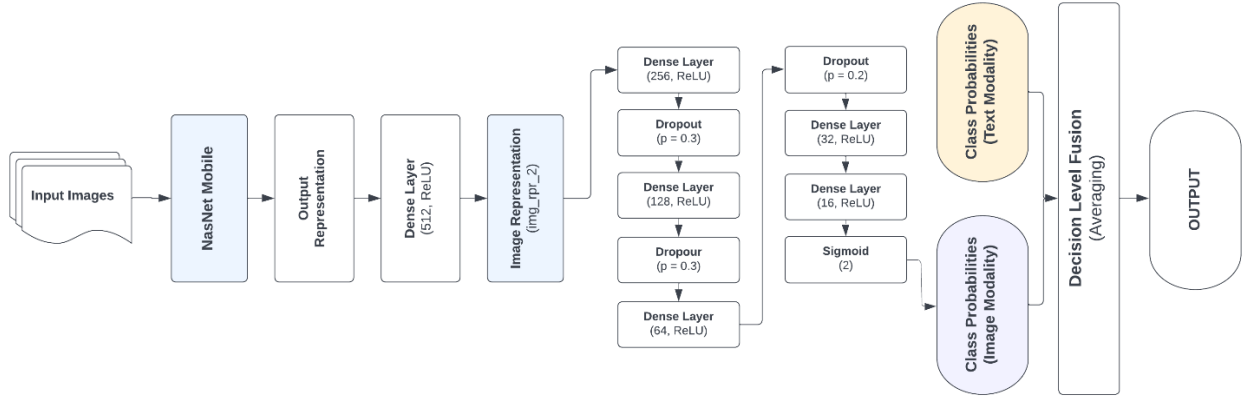

**Fig. 6.** Fusion of Modalities to arrive at final output.

### 4.5 Evaluation Metrics

The proposed method is evaluated on two datasets as mentioned before. To keep the work consistent with other counterparts, the method is evaluated on the basis of class wise precision, recall and F1 score. The overall prediction accuracy is also calculated and presented. Formulas (2-5) represent the mathematical formulation of these metrics. It should be noted that not all relevant works make use of all the metrics mentioned above. Due to this limitation, comparisons have been made with works that make use of at least two of the four mentioned metrics. Further, to evaluate the speed of the architecture, the number of parameters is also used as a point of comparison. Further, the overall multimodal training time is also provided but not compared due to a lack of publications citing the same.

$$Accuracy = \frac{TP + TN}{TP + FP + TN + FN} \tag{3}$$

$$P = \frac{TP_i}{TP_i + FP_i} \tag{4}$$

$$R = \frac{TP_i}{TP_i + FN_i} \tag{5}$$

$$F1 = \frac{2PR}{P + R} \tag{6}$$

where,
*TP* and *TN* stand for count of True Positives and True Negatives respectively,
*FP* and *FN* stand for count of False Positives and False Negatives respectively,
*P, R* and *F1* stand for Precision, Recall and F1 Score respectively,
*i* represents the $i^{th}$ class.

## 5. RESULTS AND DISCUSSION

The efficacy of the suggested framework was assessed on the datasets presented in Section 3. All neural networks were implemented using TensorFlow and the pretrained models for images and text were extracted from Keras [59] and TFHub [58] respectively. It should be noted that Keras LR was used to hyper tune the learning rate for both the neural networks shown in Fig. 5 and 6. The Adam optimizer was used on both networks with categorical cross-entropy as the loss function. All tests were run on Google Collaboratory (GPU) on a 16GB Tesla T4 GPU and 32GB of RAM. For evaluation, four metrics were used, as mentioned previously. Tables 4 and 5 present the results obtained for the two datasets and compare the same with recent relevant works, while Table 6 presents the training time results and parameter comparison. Figures 7 and 8 provide examples of generated outputs from both datasets. The section concludes by highlighting the advantages of the proposed model.

**Table 4.** Performance Comparison Summary: Twitter MediaEval Dataset

| Method | Accuracy (%) | Class: Fake (%) | | | Class: Real (%) | | |
|---|---|---|---|---|---|---|---|
| | | Precision | Recall | F1 | Precision | Recall | F1 |
| EANN [23] | 71.50 | NA | NA | NA | NA | NA | NA |
| EANN (-) [23] | 64.80 | 81.00 | 49.80 | 61.70 | 58.40 | 75.90 | 66.00 |
| MVAE [24] | 74.50 | 80.10 | 71.90 | 75.80 | 68.90 | 77.70 | 73.00 |
| SpotFake [39] | 77.80 | 75.10 | 90.00 | 82.00 | 83.20 | 60.06 | 70.10 |
| Cultural Algo [60] | 79.80 | 79.10 | 83.30 | 76.00 | 79.10 | **83.30** | 76.00 |
| Efficient Roberta [61] | 85.30 | **82.10** | 94.30 | 82.70 | 91.30 | 74.50 | 82.00 |
| **Proposed Method (SEMI-FND)** | **85.80** | 73.70 | **95.50** | **83.20** | **95.30** | 72.90 | **82.60** |

Wang et al. [23] were the first to propose an Event Adversarial Neural Network (EANN) which used an event discriminator to measure the dissimilarities among different events in order to generalize for new and emerging events. The work reported both event adversarial scores (EANN) and the event variant scores (EANN-), with the EANN achieving higher accuracy of 71.50%. However, they utilize a TextCNN [62] which is surpassed in performance by deep embeddings utilized in this work. Further, the usage of NasNet Mobile in place of VGG-19 in this paper results in the proposed model outperforming their model by 14.30%. To tackle the issue of shared representation of textual and image features, Khattar et al. [24] proposed an encoder-decoder architecture for multimodal FND (MVAE). The encoder module used by [24] included a Bi-LSTM network to encode the textual data while using a VGG-19 architecture for images. Using a decoder to train the data against the encoder, they obtained an accuracy of 74.50%. Their work, however, fell short in class wise recall as shown in Table 4. The proposed work outperforms MVAE by 11.30% while also showing a considerable increase in the class-wise recall and F1 scores. Singhal et al. [39] utilized a multimodal architecture of BERT and VGG-19 (SpotFake) with decision level fusion to achieve 77.80% accuracy. Their architecture also utilized a reduced representation of only 32 dimensions for both text and images to improve processing speed. On the other hand, the proposed framework in this work utilizes a larger dimensional representation for images and text while reducing the number of parameters by opting for smaller models. This, along with the stacked ensemble approach allows for an increase of 8% over SpotFake. In a departure from deep learning approaches, Shah et al. [60] utilized a cultural genetic algorithm to arrive at an accuracy of 79.8% which is surpassed by 6% by SEMI-FND. Finally, Singh et al. [61] used a combination of RoBERTa [63] and EfficientNetB0 [50] to improve the accuracy while reducing processing time. They reported a high accuracy of 85.30% while also reporting high F1 scores. indicating the good performance of the model on the imbalanced nature of FND. While the proposed model provides only a minor improvement of 0.5% over [61], it succeeds in significantly reducing the computational requirements as shown in Table 6.

**Table 5.** Performance Comparison Summary: Weibo Dataset

| Method | Accuracy (%) | Class: Fake (%) | | | Class: Real (%) | | |
|---|---|---|---|---|---|---|---|
| | | Precision | Recall | F1 | Precision | Recall | F1 |
| EANN | 82.70 | 82.70 | 69.70 | 75.60 | 75.20 | 86.30 | 80.40 |
| EANN (-) | 79.50 | NA | NA | NA | NA | NA | NA |
| MVAE | 82.40 | 85.40 | 76.90 | 80.90 | 80.20 | **87.50** | **83.70** |
| Efficient Roberta | 81.20 | 85.10 | 78.40 | 81.60 | 74.40 | 82.60 | 78.20 |
| **Proposed Method (SEMI-FND)** | **86.83** | **87.37** | **80.20** | **83.63** | **86.29** | 87.10 | 80.70 |

For the Weibo dataset, the proposed method is compared with four works, all of which are similar to the ones discussed in Table 4. Wang et al.'s [23] EANN architecture achieved a score of 82.70% on the Weibo dataset using their event invariant version and a score of 79.50% using the event variant one. In comparison, the proposed work achieves a score of 86.83%: an improvement of 4.13%. MVAE, proposed by Khattar et al. [24] also achieved scores similar to Wang et al. [23] with the highest reported accuracy being 82.40%. Once again, an improvement of Y% was seen when compared to SEMI-FND. Prior to the presented work, SpotFake [39] had the highest scores on the Weibo dataset with an accuracy of 89.23%. This could be attributed to the low dimensionality of the text features used by Singhal et al. [39] which reduced the errors introduced by the translated nature of the Weibo dataset (Chinese to English). While Singh et al. [61] had the highest scores on the Twitter MediaEval dataset prior to the proposed work, they obtain relatively low scores on the Weibo dataset. They attributed this to two major reasons:
1. The Weibo dataset contains a large frequency of images with human faces in comparison to the Twitter MediaEval dataset.
2. Due to the complexity of the Chinese dataset, the translation is not very exact. In addition, the posts in the Weibo dataset were longer than the shorter tweets on Twitter.

SEMI-FND, however, achieves an improvement over both these works by increasing performance over [61] by 5.63%.

**Table 6.** Comparison of Parameter Count

| Method | Image Modality | | Text Modality | |
|---|---|---|---|---|
| | Model | No. Of Parameters | Model | No. Of Parameters |
| EANN | VGG-19 | 20,024,384 | Text CNN | NA |
| MVAE | VGG-19 | 20,024,384 | Word2Vec | NA |
| SpotFake | VGG-19 | 20,024,384 | BERT Base | 110 million |
| Efficient Roberta | EfficentNet B0 | 5,300,000 | RoBERTa | 123 million |
| **SEMI-FND** | **NasNet Mobile** | **4,269,716** | **ELECTRA Small & BERT Small** | **14 + 30 = 44 million** |

While the above presented results in Table 4 and 5 show that SEMI-FND performs better than other recently proposed works, Table 6 compares the modality wise parameter count of these proposed works against that of SEMI-FND. It can be clearly seen that in the case of image modality, only [61] comes close to the low parameter count of SEMI-FND. The proposed work offers a minimum parameter reduction of 19.44% and a maximum reduction of 78.67%. Even in the text modality, the ensemble of ELECTRA and BERT reduces parameter count by 60%. This shows that the proposed work not only performs well on state-of-the-art datasets, but does so in a much more efficient and fast manner. However, none of the compared works take into account the training time metric and thus a comparison could not be made with this metric. The authors hope that this work will allow for more FND based models to provide their training times as a metric since the speed and efficiency of the model is essential in real life use cases.

From the above-presented results and discussions, the merits of the proposed method can be summarized as follows:

1. As is evident from Tables 4 and 5, the SEMI-FND model presents superior accuracy and F1 scores to notable recent works. The evaluation of the model of two datasets in different languages further reinforces its credibility and adaptability.
2. As mentioned before, the developed approach is computationally less expensive than most current works utilizing resource-heavy models such as VGG19 and LSTM-RNNs. Further, the model still maintains better accuracy than the mentioned papers.
3. As a direct consequence of the previous point, the training time of the proposed pipeline is considerably lower and offers a significant advantage for real-time applications, as early detection is especially useful in containing the spread of false information.
4. Lastly, in addition to improved accuracy, the presented framework achieves balanced scores between the real and fake classes, indicating that the model is not biased towards one particular class preventing systematic prejudice.

## 6. CONCLUSION AND FUTURE WORK

This work successfully presented an efficient multimodal framework for fake news detection utilizing BERT and ELECTRA models for extracting contextual features from given text and NasNet Mobile for processing of the associated images. It also illustrated the viability of stacking textual embeddings and decision level fusion of modalities. The proposed framework greatly reduces the compute requirements for the task along with minimizing the total time required to make the decision, while maintaining respectable accuracy. The performance of the suggested approach was evaluated using accuracy, precision, recall and F1 score on two publicly available datasets, namely Twitter and Weibo. The proposed pipeline offered an accuracy of 85.80% on the Twitter MediaEval 2015 dataset and 86.83% on the Weibo dataset. These results, being superior to their counterparts, clearly highlight the effectiveness of a multimodal approach and use of simple DNN networks in conjugation with specific processing models for each modality as opposed to conventional classifiers, adversarial networks or compute heavy DNN models such as LSTMs and RNNs.

Though the presented framework achieves superior performance over recent works, a natural progression of this work would be to incorporate user metadata in the proposed pipeline to both improve the efficacy of distinguishing between real and false information and to identify accounts that are likely to contribute to the dissemination of misinformation. The same has been difficult to accomplish due to the dearth of datasets that include resources to extract all relevant features. A promising direction would thus be to create a comprehensive and large-scale fake news benchmark dataset covering linguistic and visual features with associated user profiles and metadata. Further, leveraging concepts such as POS tagging and topic modeling would be worth investigating as they take into account textual semantics and can be used to learn the latent stance from topics, thereby for aiding models in feature extraction and fine-tuned classification. Lastly, these NLP techniques can also be combined with information retrieval and reinforcement learning as most existing research focuses on supervised learning approaches and the accuracy of semi-supervised, unsupervised, and hybrid approaches has not yet been adequately tested.